\documentclass{article}

\usepackage[final]{nips_2017}

\usepackage[utf8]{inputenc} 
\usepackage[T1]{fontenc}    
\usepackage{hyperref}       
\usepackage{url}            
\usepackage{booktabs}       
\usepackage{amsfonts}       
\usepackage{nicefrac}       
\usepackage{microtype}      
\usepackage{amsmath,graphicx}
\graphicspath{{Figures/}}
\usepackage{graphicx}
\usepackage{subcaption}
\usepackage{mwe}

\usepackage[draft]{todonotes}

\title{Cost-sensitive detection with variational autoencoders for environmental acoustic sensing}

\author{
  Yunpeng Li\\
  Department of Engineering Science\\
  University of Oxford\\
  \texttt{yli@robots.ox.ac.uk} \\
  \And
  Ivan Kiskin\\
  Department of Engineering Science\\
  University of Oxford\\
  \texttt{ikiskin@robots.ox.ac.uk} \\
  \AND
  Davide Zilli\\
  Department of Engineering Science\\
  University of Oxford\\
  \& Mind Foundry Ltd.\\
  \texttt{dzilli@robots.ox.ac.uk}
  \And
  Marianne Sinka\\
  Department of Zoology\\
  University of Oxford\\
  \texttt{marianne.sinka@zoo.ox.ac.uk} \\
  \AND
  Henry Chan\\
  Department of Engineering Science\\
  University of Oxford\\
  \texttt{tsunhenry@gmail.com} \\
  \And
  Kathy Willis\\
  Department of Zoology\\
  University of Oxford\\
  \& Royal Botanic Gardens, Kew\\
  \texttt{kathy.willis@zoo.ox.ac.uk}
  \AND
  Stephen Roberts\\
  Department of Engineering Science\\
  University of Oxford\\
  \& Mind Foundry Ltd.\\
  \texttt{sjrob@robots.ox.ac.uk}
}

\begin{document}

\maketitle

\begin{abstract}
Environmental acoustic sensing involves the retrieval
and processing of audio signals to better understand our surroundings.
While large-scale acoustic data make manual analysis infeasible,
they provide a suitable playground for machine learning approaches. Most existing machine learning techniques
developed for environmental acoustic sensing do not
provide flexible control of the trade-off
between the false positive rate and the false negative rate.
This paper presents a cost-sensitive classification
paradigm, in which the hyper-parameters of classifiers
and the structure of variational autoencoders are selected
in a principled Neyman-Pearson framework. 
We examine the performance of the proposed approach using 
a dataset from the HumBug project\footnote{humbug.ac.uk} which aims
to detect the presence of mosquitoes using sound collected
by simple embedded devices.
\end{abstract}

\section{Introduction}

Environmental acoustic sensor systems are becoming ubiquitous as we strive to improve
the understanding of our surroundings. Applications range from animal sound recognition
\cite{zilli2014,theunis2017} to smart cities~\cite{kelly2014,lloret2015}.
A significant quantity of previous work has concentrated on the development
of smartphone apps or embedded device softwares that retrieve
and transmit acoustic data. Manual analysis is usually then performed
on the collected data. However, the burden of analysis becomes unreasonable
with days, months or even years of recordings.

One popular automation approach consists of applying machine learning techniques to
detection tasks in acoustic sensing. Off-the-shelf classification 
algorithms and hand-crafted features are typically combined to complete specific
tasks~\cite{sigtia2016}. Commonly used features extracted from audio signals
include the spectrogram,  the spectral centroid, frequency-band energy features, and mel-frequency cepstral coefficient
(MFCC)~\cite{eronen2006,portelo2009,barchiesi2015}. For classification methods,
popular choices include support vector machines (SVM), hidden Markov model (HMM)-based classifiers and deep neural networks~\cite{scholler2011,lane2015,salamon2017}.

Classification metrics commonly used
in environmental acoustic sensing include
detection accuracy, the false positive and
the false negative rates~\cite{sigtia2016}.
However, there has been little research
devoted to how to obtain principled control of the false positive rate and the false negative rate for environmental acoustic sensing. The ability to control different types of errors can be important for environment sensing tasks. For example, a small false negative rate is critical in hazard event detection or rare bird species detection. Furthermore,
it would be desirable to achieve a small false positive rate if the acoustic sensor stores recordings only when it detects events of interest due to the storage limit.

In this paper, we present a principled Neyman-Pearson approach to select classifier parameters that minimise the false negative rate while keeping the false positive rate below a pre-specified threshold. This approach is similar in spirit to fusing measurements from the most informative antenna pairs for cost-sensitive microwave breast cancer detection~\cite{li2017}.
The variational autoencoder (VAE)~\cite{kingma2014} 
is used to harness the structure of the hand-crafted features
of a large amount of unlabelled data common in environmental sensing applications.
Here, we select the network structure of the VAE and the
classifier hyper-parameters automatically by minimising the Neyman-Pearson
measure~\cite{scott2007} using the ensemble
selection method~\cite{caruana2006}.
We evaluate the proposed methods using
a dataset from the HumBug project, with the aim of detecting, from audio recordings, mosquitoes
capable of vectoring malaria.

The remainder of the paper is organised
as follows:
Section~\ref{sec:method} introduces
the cost-sensitive learning
approach with a variational autoencoder. Experimental data and results are
presented in Section~\ref{sec:result}
and the conclusion is provided in
Section~\ref{sec:conclusion}.%

\section{Method}
\label{sec:method}

\subsection{Feature Extraction}

Time-frequency representations such as spectrograms
unveil important spectral characteristics of audio signals. However, their high dimensionality and
correlations between frequency contents can render
learning difficult with a small amount of data.
Cepstral coefficients, in particular mel-frequency
cepstral coefficients (MFCCs), are compact representations
of spectral envelopes that are widely used in speech
recognition and acoustic scene detection~\cite{barchiesi2015}.
In recent years, the rapid advances in both machine learning algorithms and hardware have led to very
successful applications of deep learning in speech recognition~\cite{hinton2012}. Deep learning methods
such as autoencoders have become attractive solutions
for feature extraction as the unsupervised learning
does not require training labels~\cite{blaauw2016, tan2016}.

\subsubsection{The variational autoencoder}

The variational autoencoder (VAE) is a variational inference
technique using a neural network for function approximations~\cite{kingma2014}. It has become
one of the most popular choices for unsupervised learning
of complex distributions. As a generative model,
it assumes that there is a latent variable $z \sim p_{\theta}(z)$
that influences the observation $x$ through
a conditional distribution (the probabilistic decoder) $p_{\theta}(x|z)$
parametrised by $\theta$. The variational lower bound
on the marginal likelihood of a data point $x_i$ is
\begin{align}
\log p_{\theta}(x_i) \geq L(\theta,\phi;x_i)
= -D_{KL}(q_\phi(z|x_i)||p_{\theta}(z)) + \mathbb{E}_{q_{\phi}(z|x_i)}[\log p_{\theta}(x_i|z)]\,\,,
\end{align}
where $D_{KL}$ is the KL-divergence term and the variational parameter $\phi$ specifies the recognition model (the probabilistic encoder) $q_\phi(z|x)$.
The variational autoencoder jointly optimises $\phi$ and $\theta$ with respect
to the variational lower bound $L(\theta,\phi;x_i)$.

The VAE assumes that the latent variable $z$ can be drawn from an isotropic multivariate
Gaussian distribution $p_{\theta}(z) = N(z;0,I)$ where $I$ is the identity matrix.
It is then mapped through a complex function to approximate the data generating
distribution using neural networks. More specifically, both the encoder $q_\phi(z|x)$ and the decoder $p_{\theta}(x|z)$  are modeled using multivariate Gaussian distributions with diagonal covariance matrices, where the means and variances of the Gaussian distributions
are computed using neural networks. A re-parametrisation trick is needed
to optimize the KL-divergence, by making the network differentiable so that back-propagation can be performed. We refer the readers to~\cite{kingma2014}
for more details.

\subsection{2$\nu$-SVM}

The support vector machine (SVM) is a very popular classification technique
due to its efficiency and effectiveness~\cite{cortes1995}.
It transforms the input vector $z_i$ into a high-dimensional space through a mapping function
$h()$. The intuition is that
the separation of two classes is easier in this
transformed high-dimensional space in which the SVM
constructs a max-margin classifier.
The classification score $f(z)$ is defined as $f(z) = w^T h(z)+b$,
where $w$ is the normal vector to the decision hyperplane
and $b$ is the bias term that shifts the hyperplane.
We can avoid explicitly evaluating
$h()$ using a kernel trick.
Slack variables $\epsilon_i \geq 0$ are introduced
as in general the two classes cannot be separated
even in the high-dimensional space. A value $\epsilon_i > 0$ indicates a margin error that
the data point $z_i$ lies on the wrong side of the decision hyperplane.
The SVM maximizes the margin while penalizing margin error. Popular variants of the SVM include the $C$-SVM~\cite{cortes1995} and the $\nu$-SVM~\cite{scholkopf2000}.


For the $\nu$-SVM, the maximum margin solution is
a quadratic programming problem:
\begin{align}
\min_{w,b,\epsilon,\rho}&{\dfrac{1}{2}||w||^2 -\nu\rho+ \dfrac{1}{n}\sum_{i=1}^n\epsilon_i}\\
\mbox{subject to } \epsilon_i &\geq 0, \rho \geq 0,
y_{i}f(z_i)\geq \rho-\epsilon_i, \forall i \nonumber \,\,.
\end{align}
This formulation allows for a straightforward interpretation of the 
parameters in the minimization. The parameter $\nu \in [0,1]$ serves as an upper
bound on the fraction of margin errors and a lower bound on the
fraction of support vectors~\cite{scholkopf2000}. The parameter $\rho$ influences the width of the margin. $n$ is the number of data points.

To allow for cost-sensitive classification, Chew et al.~proposed the 2$\nu$-SVM
by introducing an additional parameter to produce
an asymmetric error~\cite{chew2001,davenport2010}.
\begin{align}
&\min_{w,b,\epsilon,\rho}{\dfrac{1}{2}||w||^2 -\nu\rho+ \dfrac{w_+}{n}\sum_{i\in i_+}\epsilon_i + \dfrac{1-w_+}{n}\sum_{i\in I_-}\epsilon_i}\\
&\mbox{subject to } \epsilon_i \geq 0, \rho \geq 0,
y_if(z_i)\geq \rho-\epsilon_i, \forall i \nonumber.
\end{align}
$I_{+}$ denotes the set of data elements with the label $y_i=+1$, and
$I_{-}$ denotes the set of data elements with
the label $y_i=-1$. 
We can express the problem in a different way by introducing
parameters $\nu_+\in [0, 1]$ and $\nu_-\in [0, 1]$
to replace $\nu$ and $w_+$ (hence the name $2\nu$-SVM).
$\nu_+$ and $\nu_-$ bound the fractions of
margin errors and support vectors from each class~\cite{davenport2010}.

\subsection{Cost-sensitive ensemble selection}

In order to perform cost-sensitive classification, we need
an objective function to gauge the performance of a classifier
with different cost constraints. Scott et al.\ proposed a
scalar performance measure $\hat{e}$ in~\cite{scott2007}
that soft-constrained the false positive rate of the classifier
to be below a target value $\alpha$, while minimising the false negative rate:
\begin{align}
\hat{e} = \dfrac{1}{\alpha}\max\{\hat{P}_F-\alpha,0\}+\hat{P}_M\,\,
\label{eq:NP_measure_empirical}
\end{align}
where $\widehat{P}_F$ and $\widehat{P}_M$ are the empirical false positive rate
and the empirical false negative rate, respectively.

\begin{figure}[htbp]
\centering
\includegraphics[scale=0.6]{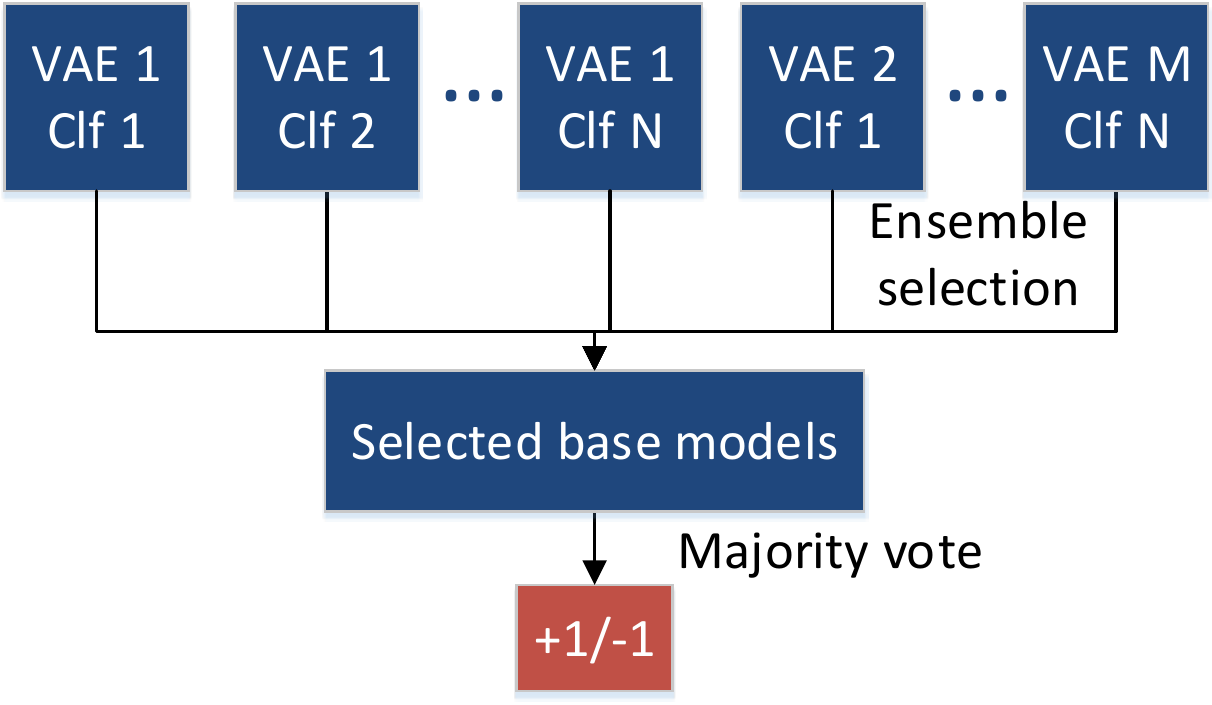}
\caption{The ensemble selection approach. There are $M$ different VAE network
candidate structures and $N$ different combinations of classifier hyper-parameter values.}
\label{fig:ensemble_selection}
\end{figure}

We may expect that certain variational autoencoder configurations are more effective
in representation learning for specific audio signals. Different classifier
hyper-parameters, e.g. $\nu_+$, $\nu_-$ and the detection threshold
of detector outputs, may suit different cost objectives to
varying degrees. We can apply the ensemble selection framework proposed in~\cite{caruana2006}
to form a model ensemble that constitutes the most informative base models
which minimise the Neyman-Pearson measure $\hat{e}$ in~\eqref{eq:NP_measure_empirical}.
In our context, base models are models with different autoencoder network structures
and classifier hyper-parameters. Note that the base classifiers are not restricted to
the $2\nu$-SVM in the cost-sensitive ensemble selection framework. Any classifier with probabilistic outputs can be adopted, as
different types of errors can be controlled
through a detection threshold applied on the probabilistic output.
After the model selection, the classification
decision in the test stage will be a majority vote among the committee of the selected $Q$ best models.
The ensemble selection architecture is shown in Figure~\ref{fig:ensemble_selection}.
 
\section{Data and Results}
\label{sec:result}

\subsection{Dataset}

%
%
%

We conducted experiments using a dataset collected
in the HumBug project~\footnote{see humbug.ac.uk}.
The HumBug project aims to detect malaria-vectoring mosquitoes through environment sound.
The dataset used here includes 57 audio recordings
with a total length of around 50 minutes.
736 seconds of these recordings contain sound of the Culex quinquefasciatus mosquitoes.
We split these recordings into short audio clips, or what we call samples, with a duration of 0.1 seconds.
Labels are given to each of these short audio clips.
We resample to obtain a balanced dataset. Hence the dataset contains
$7360$ positive samples (with mosquito sound) and $7360$ negative
samples (no mosquito sound).
A random sampling approach~\cite{li2017e},
which randomly samples audio clips without replacement
in the data set, was used to form the trainings set
with $10\%$ of total samples. The remaining $90\%$ samples
are used for testing.

\subsection{Parameter values}

Our target false positive rate threshold is set to $0.1$, i.e. we would like to
minimise the missed detection while maintaining the false alarm rate
to be below $10\%$. The SVM with the RBF kernel and the MFCC feature
serves as the benchmark algorithm,
as it leads to the best performance among
a dozen of audio features (spectrogram, specentropy, etc.) and traditional classifiers (random forest, naive Bayes, etc.).

Candidate parameter values used to form the model library include
the $2\nu$-SVM parameters: $\gamma: 2^{-5}, 2^{-3}, \ldots, 2^{5}$ and
	$\nu_+/ \nu_-: 10^{-5}, 3\times10^{-5}, 10^{-4}, 3\times10^{-4},0.001, 0.003, 0.01, 0.03, 0.1,$ $  0.2, 0.3,0.4, \ldots, 1$.
The MFCC feature is 13-dimensional. To reduce feature dimension while
maintaining most of detection power, we would like to use the VAE for feature re-representation.
The candidate dimensions of the latent variable of the VAE include $3$ and $5$, while the number of nodes in the hidden layer can be either $10$ or $50$. Note that these choices are representative only and can vary for different datasets.

We initialise parameters of the VAE using a normal distribution with a standard deviation $0.01$.
Adam~\cite{kingma2014b} is used in the optimisation of the VAE.
The cost-sensitive detector forms an ensemble of
100 base models to produce the final prediction.

\subsection{Results}

\begin{figure}[htbp]
\centering
\begin{subfigure}[b]{0.32\textwidth}
    \centering
    \includegraphics[width=1.3\textwidth]{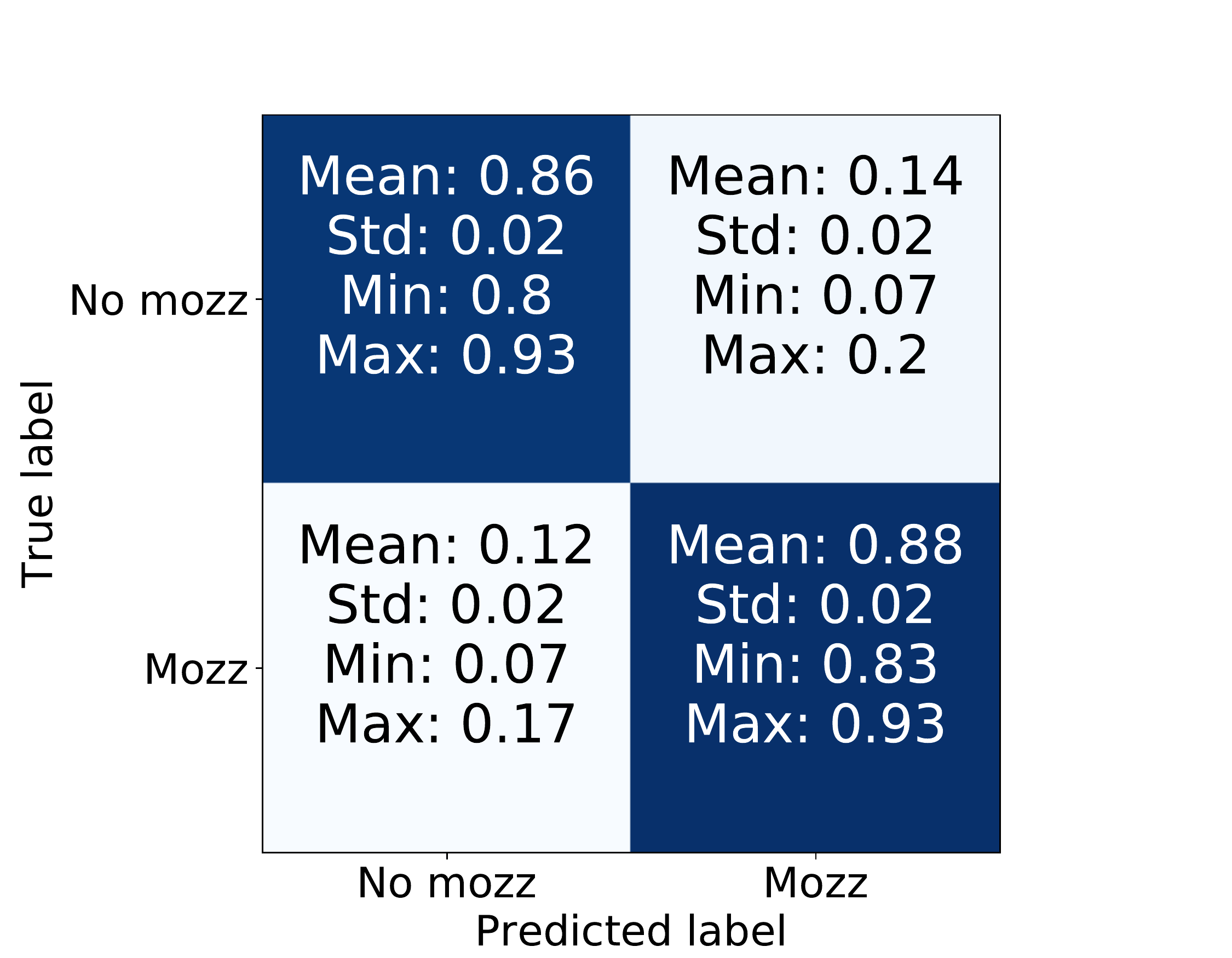}
    \caption[]%
    {{\small MFCC, SVM}}    
    \label{fig:MFCC_SVM}
\end{subfigure}
\hfill
\begin{subfigure}[b]{0.32\textwidth}  
    \centering 
    \includegraphics[width=1.3\textwidth]{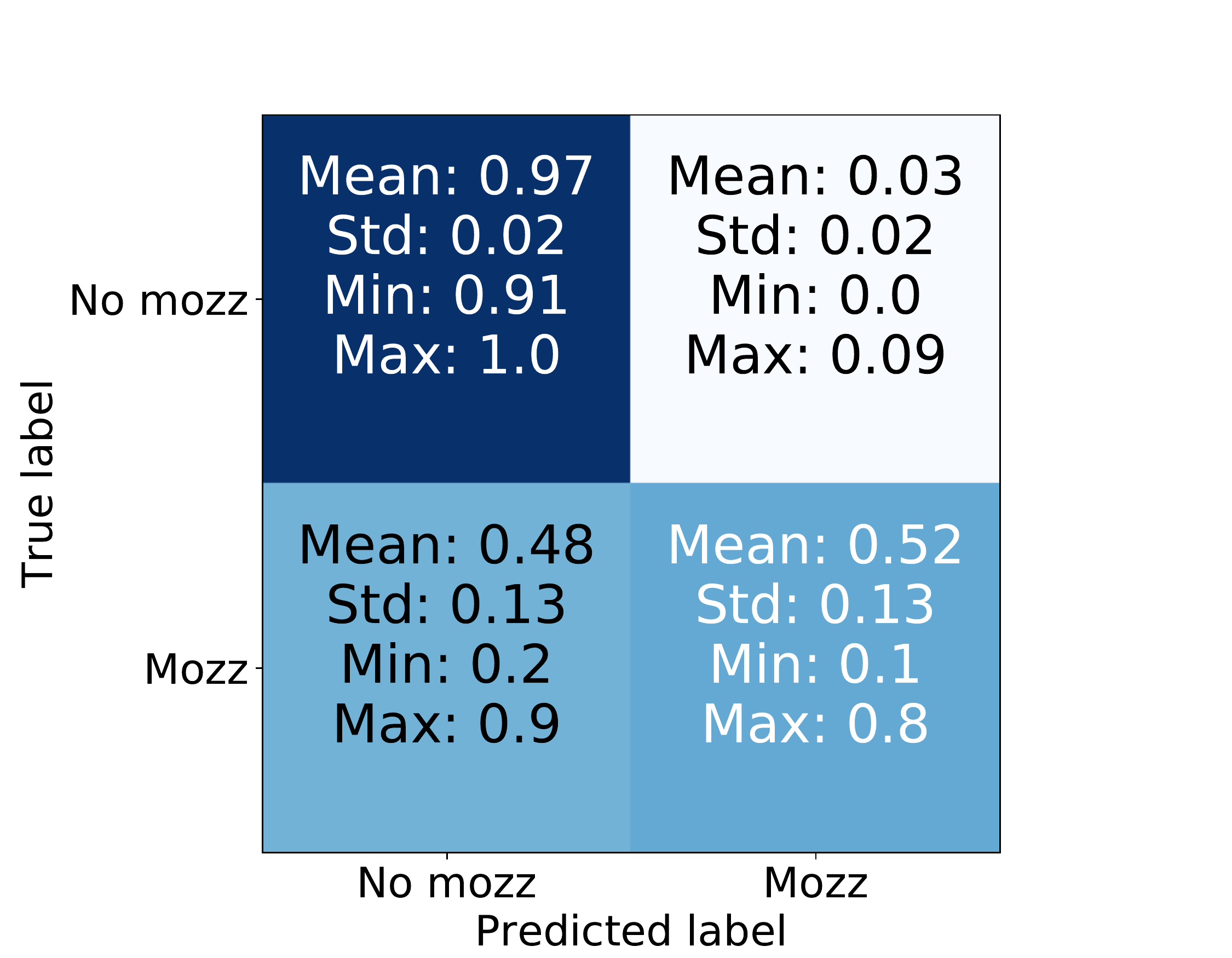}
    \caption[]%
    {{\small MFCC, CSSVM}}    
    \label{fig:MFCC_CSSVM}
\end{subfigure}
\hfill
\begin{subfigure}[b]{0.32\textwidth}   
    \centering 
    \includegraphics[width=1.3\textwidth]{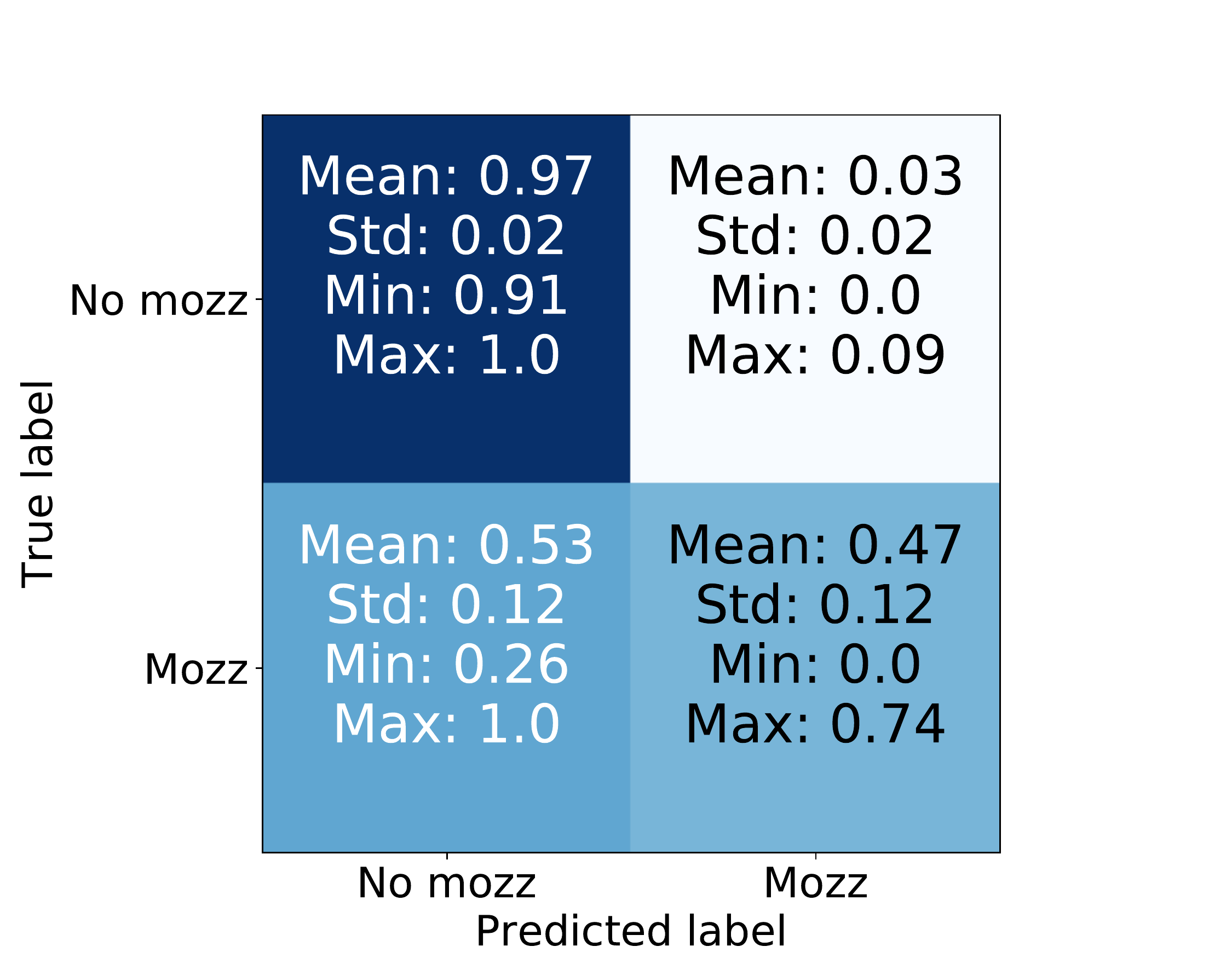}
    \caption[]%
    {{\small MFCC+VAE, CSSVM}}    
    \label{fig:MFCC_VAE_CSSVM}
\end{subfigure}
\caption[]
{\small Confusion matrices of the test errors with different features and classifiers. ``Mozz'' is the positive class.} 
\label{fig:result}
\end{figure}

We performed 100 simulation trials, in which each trial differs due to the random seed initialisation,
hence producing different data partitions and initial parameter values. 
We see from Figure~\ref{fig:MFCC_CSSVM} and Figure~\ref{fig:MFCC_VAE_CSSVM} that the cost-sensitive SVM (CSSVM) framework
is able to maintain the false positive rate below $0.1$ in all simulation
trials. The SVM with the MFCC feature has impressive performance (Figure~\ref{fig:MFCC_SVM}), but it fails to control the false positive rate to be below our target value.

Although the VAE leads to slightly smaller sensitivity
in detection performance for this dataset, the VAE provides a flexible mechanism
to reduce feature dimension in the ensemble model.
The reduction in the model size can be attractive in porting the model into embedded devices for environmental acoustic sensing~\cite{li2017e}.

\section{Conclusion}
\label{sec:conclusion}

This paper presents a cost-sensitive classification framework for
environmental audio sensing. The ensemble selection techniques
are able to choose hyper-parameters of the feature extraction methods
and the classifier in a principled manner. The proposed cost-sensitive
SVM framework with the MFCC features is shown to ensure that
the false positive rate lies below a pre-specified target value
while minimising the false negative rate, by selecting a committee of
best-performing individual models in a model library containing thousands of base models.
We also consider VAE feature re-representations, which are helpful in selecting simple feature representations for low-power embedded systems.

\subsubsection*{Acknowledgements}

This work is part-funded by a Google Impact Challenge
award.


\small

\bibliographystyle{apalike}
\bibliography{humbug}

\begin{thebibliography}{}

\bibitem[Barchiesi et~al., 2015]{barchiesi2015}
Barchiesi, D., Giannoulis, D., Stowell, D., and Plumbley, M.~D. (2015).
\newblock Acoustic scene classification: Classifying environments from the
  sounds they produce.
\newblock {\em IEEE Signal Processing Mag.}, 32(3):16--34.

\bibitem[Blaauw and Bonada, 2016]{blaauw2016}
Blaauw, M. and Bonada, J. (2016).
\newblock Modeling and transforming speech using variational autoencoders.
\newblock In {\em Proc. Interspeech}, pages 1770--1774.

\bibitem[Caruana et~al., 2006]{caruana2006}
Caruana, R., Munson, A., and Niculescu-Mizil, A. (2006).
\newblock Getting the most out of ensemble selection.
\newblock In {\em Proc. Int. Conf. Data Mining (ICDM)}, pages 828--833.

\bibitem[Chew et~al., 2001]{chew2001}
Chew, H.-G., Bogner, R.~E., and Lim, C.-C. (2001).
\newblock Dual $\nu$-support vector machine with error rate and training size
  biasing.
\newblock In {\em Proc. Int. Conf. Acoustics, Speech and Signal Proc.
  (ICASSP)}, pages 1269--1272, Salt Lake City, USA.

\bibitem[Cortes and Vapnik, 1995]{cortes1995}
Cortes, C. and Vapnik, V. (1995).
\newblock Support-vector networks.
\newblock {\em Mach. Learn.}, 20(3):273--297.

\bibitem[Davenport et~al., 2010]{davenport2010}
Davenport, M.~A., Baraniuk, R.~G., and Scott, C.~D. (2010).
\newblock Tuning support vector machines for minimax and {N}eyman-{P}earson
  classification.
\newblock {\em IEEE Trans. Pattern Anal. Mach. Intell.}, 32(10):1888--1898.

\bibitem[Eronen et~al., 2006]{eronen2006}
Eronen, A.~J., Peltonen, V.~T., Tuomi, J.~T., Klapuri, A.~P., Fagerlund, S.,
  Sorsa, T., Lorho, G., and Huopaniemi, J. (2006).
\newblock Audio-based context recognition.
\newblock {\em IEEE/ACM Trans. Speech Audio Process.}, 14:321--329.

\bibitem[Hinton et~al., 2012]{hinton2012}
Hinton, G., Deng, L., Yu, D., Dahl, G.~E., r.~Mohamed, A., Jaitly, N., Senior,
  A., Vanhoucke, V., Nguyen, P., Sainath, T.~N., and Kingsbury, B. (2012).
\newblock Deep neural networks for acoustic modeling in speech recognition: The
  shared views of four research groups.
\newblock {\em IEEE Signal Processing Mag.}, 29(6):82--97.

\bibitem[Kelly et~al., 2014]{kelly2014}
Kelly, B., Hollosi, D., Cousin, P., Leal, S., Iglár, B., and Cavallaro, A.
  (2014).
\newblock Application of acoustic sensing technology for improving building
  energy efficiency.
\newblock {\em Procedia Computer Science}, 32:661 -- 664.

\bibitem[Kingma and Ba, 2014]{kingma2014b}
Kingma, D. and Ba, J. (2014).
\newblock Adam: A method for stochastic optimization.
\newblock {\em arXiv:1412.6980}.

\bibitem[Kingma and Welling, 2014]{kingma2014}
Kingma, D. and Welling, M. (2014).
\newblock Auto-encoding variational {B}ayes.
\newblock In {\em Proc. Intl. Conf. Learning Representations (ICLR)}.

\bibitem[Lane et~al., 2015]{lane2015}
Lane, N.~D., Georgiev, P., and Qendro, L. (2015).
\newblock Deepear: Robust smartphone audio sensing in unconstrained acoustic
  environments using deep learning.
\newblock In {\em Proc. ACM Intl. J. Conf. Pervasive Ubiquitous Computing
  (UbiComp)}, pages 283--294.

\bibitem[Li et~al., 2017a]{li2017}
Li, Y., Porter, E., Santorelli, A., Popovi\'c, M., and Coates, M. (2017a).
\newblock Microwave breast cancer detection via cost-sensitive ensemble
  classifiers: Phantom and patient investigation.
\newblock {\em Biomed. Signal Process. Control}, 31:366 -- 376.

\bibitem[Li et~al., 2017b]{li2017e}
Li, Y., Zilli, D., Chan, H., Kiskin, I., Sinka, M., Roberts, S., and Willis, K.
  (2017b).
\newblock Mosquito detection with low-cost smartphones: data acquisition for
  malaria research.
\newblock In {\em NIPS Workshop on Machine Learning for the Developing World},
  Long Beach, USA.
\newblock arXiv:1711.06346.

\bibitem[Lloret et~al., 2015]{lloret2015}
Lloret, J., Canovas, A., Sendra, S., and Parra, L. (2015).
\newblock A smart communication architecture for ambient assisted living.
\newblock {\em IEEE Commun. Mag.}, 53:26--33.

\bibitem[Portelo et~al., 2009]{portelo2009}
Portelo, J., Bugalho, M., Trancoso, I., Neto, J., Abad, A., and Serralheiro, A.
  (2009).
\newblock Non-speech audio event detection.
\newblock In {\em Proc. Intl. Conf. Acoustics, Speech and Signal Proc.
  (ICASSP)}, pages 1973--1976.

\bibitem[Salamon and Bello, 2017]{salamon2017}
Salamon, J. and Bello, J.~P. (2017).
\newblock Deep convolutional neural networks and data augmentation for
  environmental sound classification.
\newblock {\em IEEE Signal Process. Lett.}, 24(3):279--283.

\bibitem[Sch\"{o}lkopf et~al., 2000]{scholkopf2000}
Sch\"{o}lkopf, B., Smola, A.~J., Williamson, R.~C., and Bartlett, P.~L. (2000).
\newblock New support vector algorithms.
\newblock {\em Neural Comput.}, 12(5):1207--1245.

\bibitem[Scholler and Purwins, 2011]{scholler2011}
Scholler, S. and Purwins, H. (2011).
\newblock Sparse approximations for drum sound classification.
\newblock {\em IEEE J. Sel. Topics Signal Process.}, 5(5):933--940.

\bibitem[Scott, 2007]{scott2007}
Scott, C. (2007).
\newblock Performance measures for neyman-pearson classification.
\newblock {\em IEEE Trans. Inf. Theory}, 53:2852--2863.

\bibitem[Sigtia et~al., 2016]{sigtia2016}
Sigtia, S., Stark, A.~M., Krstulović, S., and Plumbley, M.~D. (2016).
\newblock Automatic environmental sound recognition: Performance versus
  computational cost.
\newblock {\em IEEE/ACM Trans. Speech Audio Process.}, 24:2096--2107.

\bibitem[Tan and Sim, 2016]{tan2016}
Tan, S. and Sim, K.~C. (2016).
\newblock Learning utterance-level normalisation using variational autoencoders
  for robust automatic speech recognition.
\newblock In {\em Proc. IEEE Spoken Language Technology Workshop (SLT)}, pages
  43--49.

\bibitem[Theunis et~al., 2017]{theunis2017}
Theunis, J., Stevens, M., and Botteldooren, D. (2017).
\newblock Sensing the environment.
\newblock In {\em Participatory Sensing, Opinions and Collective Awareness},
  pages 21--46. Springer International Publishing, Switzerland.

\bibitem[Zilli et~al., 2014]{zilli2014}
Zilli, D., Parson, O., Merrett, G.~V., and Rogers, A. (2014).
\newblock A hidden {M}arkov model-based acoustic cicada detector for
  crowdsourced smartphone biodiversity monitoring.
\newblock {\em J. Artif. Intell. Res.}, 51:805--827.

\end{thebibliography}

\end{document}